\documentclass[11pt]{article}
\usepackage[hyperref]{ccl2022-en}
\usepackage{times}
\usepackage{url}
\usepackage{amsmath}
\usepackage{latexsym}
\usepackage{fancyhdr}
\usepackage{amssymb}
\usepackage{booktabs}

\usepackage{graphicx}
\usepackage[utf8]{inputenc}
\usepackage{CJKutf8}

\title{A Two-Stage Method for Chinese AMR Parsing}

\author{
 Liang Chen, \ Bofei Gao, \   Baobao Chang
\\ 
Key Laboratory of Computational Linguistics, Peking University, MOE, China \\
 \texttt{leo.liang.chen@outlook.com} \\
 \texttt{gaobofei@163.com} \\
 \texttt{chbb@pku.edu.cn}
}

\date{}

\begin{document}
\maketitle
\begin{abstract}
In this paper, we provide a detailed description of our system at CAMRP-2022 evaluation. We firstly propose a two-stage method to conduct Chinese AMR Parsing with alignment generation, which includes Concept-Prediction and Relation-Prediction stages. Our model achieves 0.7756 and 0.7074 Align-Smatch F1 scores on the CAMR 2.0 test set and the blind-test set of CAMRP-2022 individually. We also analyze the result and the limitation such as the error propagation and class imbalance problem we conclude in the current method. Code and the trained models are released at  \href{https://github.com/PKUnlp-icler/Two-Stage-CAMRP}{https://github.com/PKUnlp-icler/Two-Stage-CAMRP} for reproduction. 
\end{abstract}

\section{Introduction}
\label{intro}

%
% The following footnote without marker is needed for the camera-ready
% version of the paper.
% Comment out the instructions (first text) and uncomment the 8 lines
% under "final paper" for your variant of English.
%
% \cclfootnote{
%     %
%     % for review submission
%     %
%     \hspace{-0.65cm}  % space normally used by the marker
%     % Place licence statement here for the camera-ready version. See Section~\ref{licence} of the instructions for preparing a manuscript.
%     \textcopyright 2022 China National Conference on Computational Linguistics

%     \noindent Published under Creative Commons Attribution 4.0 International License
% }

\begin{figure}[h]
    \centering
    \includegraphics[width=0.7\linewidth]{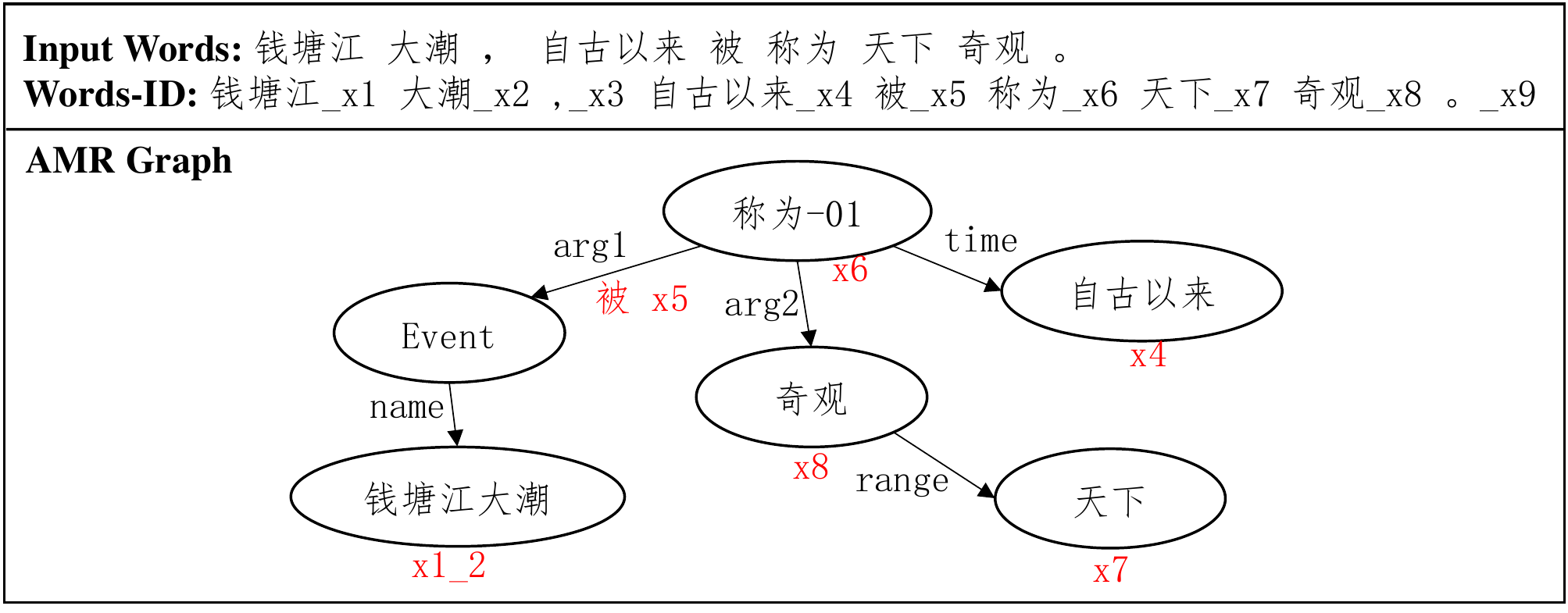}
    \caption{An example of Chinese AMR. Red word-IDs under concepts denote concept alignment. Red words and word-IDs under relation denote relation alignment.  }
    \label{fig:camr}
\end{figure}

Abstract Meaning Representation (AMR) \cite{ban-AMR} parsing targets to transform a sentence into a directed acyclic graph, which represents the relations among different concepts. The original AMR does not provide concept-to-word alignment information, which hinders the trace-back from concept to input word and brings difficulties to AMR parsing. To solve the problem, based on Chinese AMR \cite{li-etal-2016-annotating} , \newcite{Li_Wen_Song_Qu_Xue_2019} further propose to add concept and relation alignment to the structure of Chinese AMR as shown in Figure~\ref{fig:camr}.

Currently, while a majority of work is focusing on improving the performance of English AMR Parsing \cite{xu-seqpretrain,bevil-spring,HCL,bai-etal-2022-graph,chen-etal-2022-atp,drozdov-etal-2022-inducing}, those methods or models can not be directly applied to Chinese AMR Parsing since English AMR does not provide alignment information itself. To better reflect the full structure of Chinese AMR, CAMRP-2022 evaluation\footnote{https://github.com/GoThereGit/Chinese-AMR} firstly requires the AMR parser to generate explicit word alignment including concept and relation alignment which calls for novel models and algorithms.

We propose a two-stage method to conduct Chinese AMR Parsing with alignment generation\footnote{We participate in the closed-track evaluation where we can only use HIT-roberta\cite{cui-etal-2020-revisiting} as the pretrained language model}. In a nutshell, the method includes the Concept-Prediction and Relation-Prediction stages, which can be regarded as the process of graph formation. In the Concept-Prediction stage, we develop a hierarchical sequence tagging framework to deal with the concept generation and the complex multi-type concept alignment problem. In the Relation-Prediction stage, we utilize the biaffine network to predict relations and Relation-alignment simultaneously among predicted concepts. Our model ranks 2nd in the closed-track of the evaluation, achieving 0.7756 and 0.7074 Align-Smatch \cite{alignsmatch} F1 scores on the CAMR 2.0 test set and the blind-test set of CAMRP-2022 individually.

\section{Method}

\begin{figure}[h]
    \centering
    \includegraphics[width=1\linewidth]{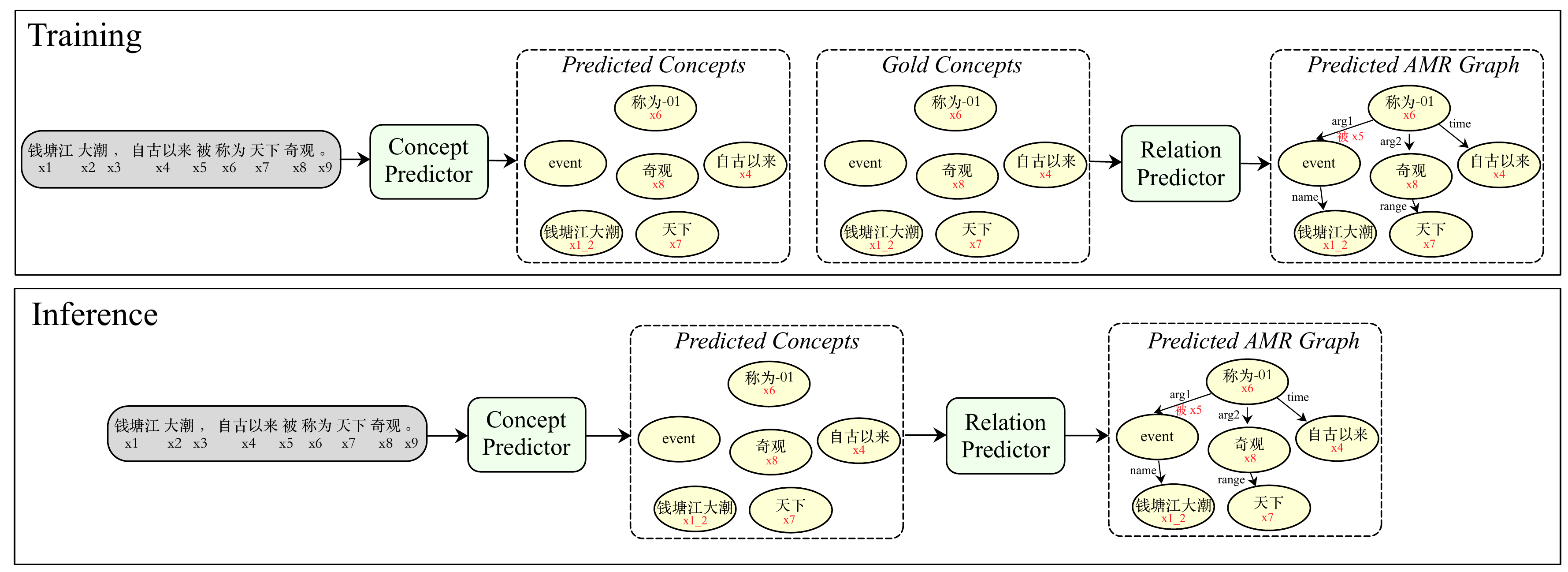}
    \caption{Pipeline for the two-stage method. In the training phase, the relation prediction stage takes the concepts from the gold AMR graph as input. During the inference phase, the relation prediction stage takes the output of the concept predictor as input.}
    \label{fig:piplines}
\end{figure}

Our methods includes the Concept-Prediction stage and Relation-Prediction stage. As illustrated in Figure~\ref{fig:piplines}, during training, both stages have individual input and output. During inference, the output concepts from the Concept-Prediction stage are passed to the Concept-Prediction stage to generate the full AMR graph.

\subsection{Concept-Prediction}

Different from English AMR where nodes or concepts can have arbitrary variable names, a large portion of concepts of Chinese AMR have standard variable names which denote the alignment to input words. Moreover, there are different alignment patterns which make generating the right alignment a complex problem. In following sections, we'll sequentially introduce the alignment rules we design for CAMRP-2022, the two-stage method and the model.

\subsubsection{Multi-Type Concept Alignment Rule}
\begin{figure}[t]
    \centering
    \includegraphics[width=1\linewidth]{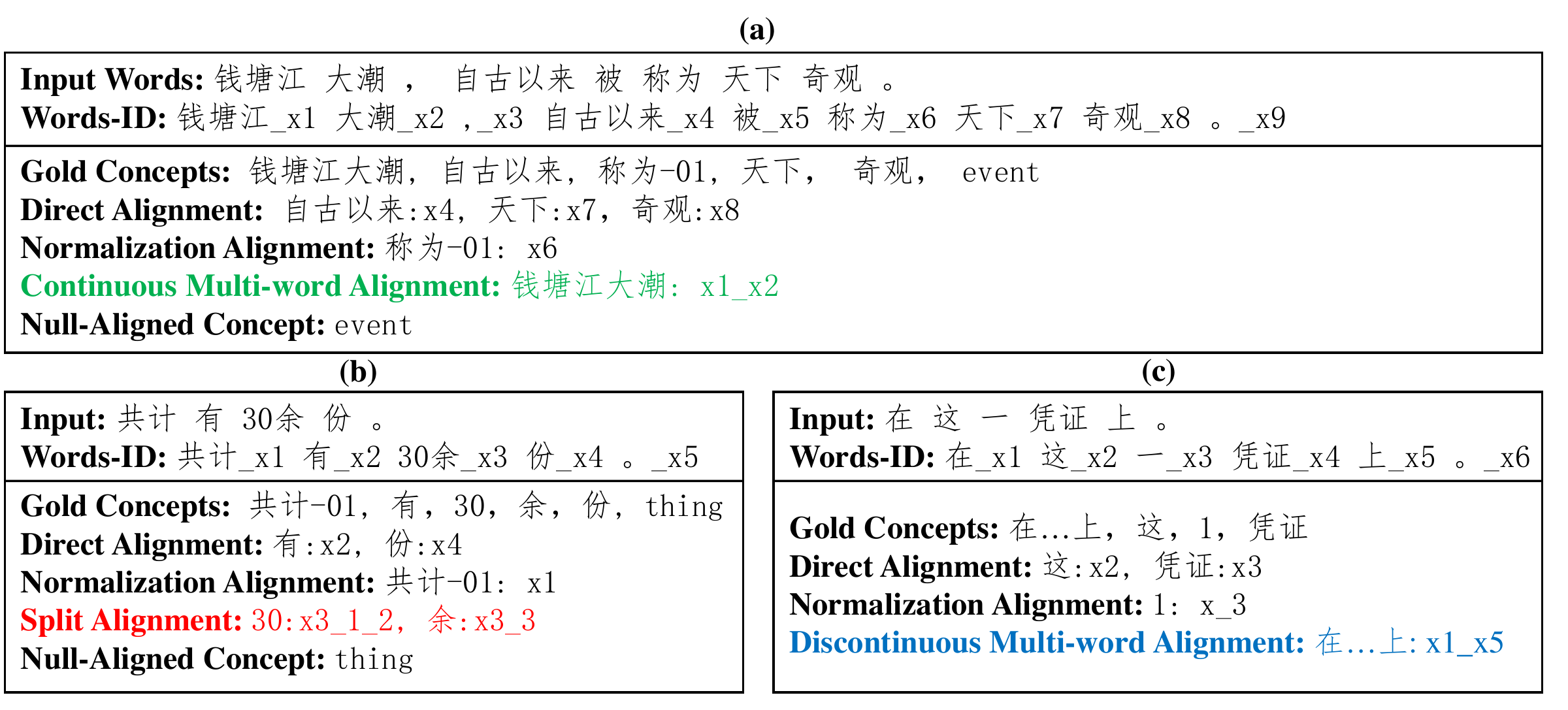}
    \caption{Example of different Concept Alignment cases. Gold concepts denote all concepts in the gold AMR graph of the input sentence. The concepts can be divided into different categories according to the alignment rules. We use words in color to represent the unique alignments in each example.}
    \label{fig:concept_align}
\end{figure}

We mainly design 6 different alignment rules for concepts according to different alignment patterns, which are \textbf{\textit{Direct Alignment}}, \textbf{\textit{Normalization Alignment}}, \textbf{\textit{Continuous Multi-word Alignment}}, \textbf{\textit{Discontinuous Multi-word Alignment}}, \textbf{\textit{Split Alignment}} and \textbf{\textit{Null-Aligned Concepts}}. The difference among alignment rules lies in how an abstract concept corresponds to the input words. We list three cases involving different rules as shown in Figure~\ref{fig:concept_align} as examples.

\begin{enumerate}

    \item \textbf{Direct Alignment} is the easiest alignment where a concept directly corresponds to a certain word in the input without the need for any modification. 
    \item \textbf{Normalization Alignment} exists when a concept still corresponds to one word in the input however needs to be ``normalized'' into the final concept. The normalization includes different situations like word sense disambiguation for predicate and Arabic numerals transformation for numerals in other languages. For example, as shown in Figure~\ref{fig:concept_align}, in case (a) the word ``\begin{CJK*}{UTF8}{gbsn}称为\end{CJK*}'' corresponds to the concept  ``\begin{CJK*}{UTF8}{gbsn}称为-01\end{CJK*}'' after word sense disambiguation. In case (c), Chinese numeral ``\begin{CJK*}{UTF8}{gbsn}一\end{CJK*}'' would be mapped to concept ``1'' since all numeral concepts in  CAMR are Arabic.  
    \item \textbf{Continuous Multi-word Alignment} exists when multiple continued words in the input sentence are concatenated into the final concept, which usually happens for named entities. 
    \item \textbf{Discontinuous Multi-word Alignment} means multiple discontinued words in the input sentence are joined into the final concept or preposition patterns like ``\begin{CJK*}{UTF8}{gbsn}在...上\end{CJK*}'' .
    \item \textbf{Split Alignment} denotes one word that could correspond to multiple concepts, which usually suggests the word corresponds to a sub-graph in the final AMR graph. 
    \item \textbf{Null-Aligned Concepts} do not have alignment and could have arbitrary variable names. These concepts usually abstract away from syntactic features and do not directly correspond to certain word in the input sentence, making them harder for the model to predict. In fact, according to our experiment, our system could reach a 0.91 f1 score for aligned concepts' prediction but only a 0.70 f1 score for Null-Aligned concepts' prediction.

\end{enumerate}

\begin{figure}[b]
    \centering    \includegraphics[width=1\linewidth]{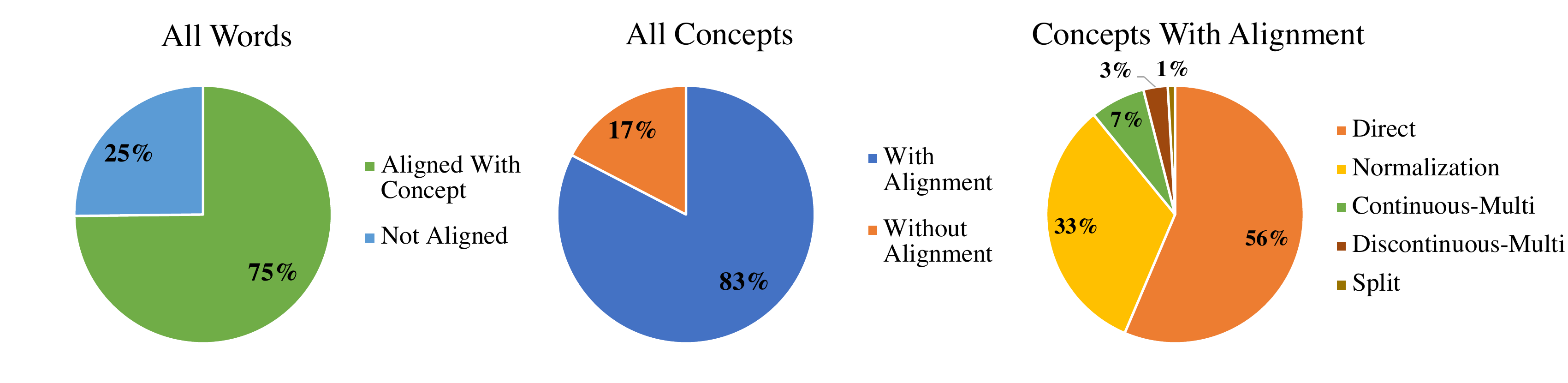}
    \caption{Statistics about the alignment between concepts and input words in CAMR 2.0.}
    \label{fig:concept_align_pie}
\end{figure}

As shown in Figure \ref{fig:concept_align_pie}, we further collect more statistics about the alignment between concepts and input words with the training set of the CAMR 2.0 dataset. From the perspective of input sentences, about 75\% of words in the input sentences are associated with certain concepts under one alignment rule. From the perspective of concepts, there are 83\% of concepts with alignment. For all concepts with alignment, a majority of them belongs to Direct(56\%) and Normalization(33\%) Alignments.

\subsubsection{Sequence Tagging Framework}

\begin{figure}[t]
    \centering
    \includegraphics[width=1\linewidth]{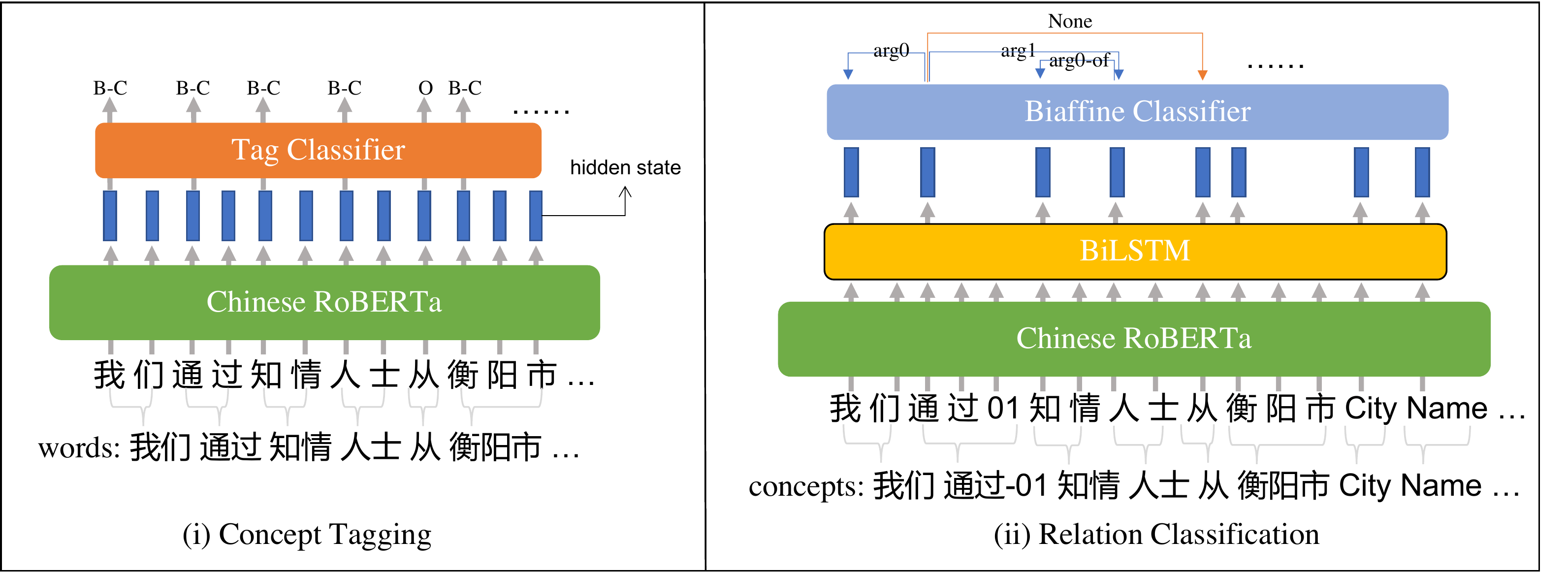}
    \caption{The Two-Stage Parsing Model. We use the same Concept Tagging model structure to conduct three hierarchical sequence tagging tasks, each with a different Tag Classifier. Relation Classification takes concepts as input and the Biaffine Classifier outputs the relation between every two concepts. In both models, words or concepts are first splitted into characters before feeding into the pretrained language model and we use the hidden state of the first character to represent the word in the last classifier layers.  }
    \label{fig:models}
\end{figure}

In spite of the complex word alignment rules, we can see that a large portion of concepts are directly or indirectly aligned with a single word of input and one word can only correspond to one concept at most, which inspires us to adopt sequence tagging method. It can deal with concept prediction and Direct Alignment prediction simultaneously. Considering different alignment rules, we develop three sequence tagging rules to cover all possible situations.

\paragraph{Model Structure}

As depicted in Figure~\ref{fig:models}, we add a linear layer on the top of the Chinese RoBERTa model as a Tag Classifier. Adapting to character-based Chinese pretrained language model, Tag classification is conducted \textbf{on the first character's hidden state} for a word with multiple characters. During training, we use Cross-Entropy as the loss function and use the average loss of  $N$ all tags in a sentence as the final loss, as described in Equation~\ref{eq:tagging},

\begin{equation}
\begin{aligned}
    \text{TagCLS}(\mathbf{a})  & = [\mathbf{a};\mathbf{1}]\mathbf{W}, (\mathbf{a} \in \mathbb{R}^{1 \times d},  \mathbf{W} \in \mathbb{R}^{(d+1)\times c} ) \\
    \text{Loss}&=\frac{1}{N}\sum_{i=1}^N  CE(\text{TagCLS}(\mathbf{h_i}) ,\hat{\mathbf{h_i}})
\end{aligned}
\label{eq:tagging}
\end{equation}

where $d$ denotes the hidden size, $c$ denotes the number of different tags, $N$ denotes the number of input words, $\mathbf{h_i}$ denotes the $i^{th}$ word's output hidden state from the encoder and $\hat{\mathbf{h_i}}$ denotes the one-hot vector of its gold tag.

\paragraph{Surface Tagging}
We design an 8-classes BIO tagging rule as the first step to process the input sentence. The eight classes are O, B-Single, B-Continuous-Multiword, I-Continuous-Multiword, B-Discontinuous-Multiword, I-Discontinuous-Multiword, B-Split, and B-Virtual. This tagging rule can cover 4 out of 6 alignment rules, which are Direct Alignment, Continuous Multi-word Alignment, Discontinuous Multi-word Alignment, and Split Alignment. Note that the B-Single tag is for both Direct Alignment and Normalization Alignment because they both correspond to one input word. As for B-Split, we use manually curated rules to split the word with Split Alignment. Note that B-Virtual is also added to label the virtual word for the later relation classification task.  The F1-score of the Surface Tagging step can reach 91\% on the development set in our experiment.

\paragraph{Normalization Alignment Tagging} Previous Surface Tagging can not recognize words that need normalization like word sense disambiguation so we introduce a 2-class Tagging rule to identify whether a word from the input sentence needs normalization before becoming a concept in the AMR graph. The labels can be collected directly from the gold AMR graph. If one concept is aligned to one identical word from the input sentence, then the word's label is negative. If the concept is aligned to a word different from itself, then the word's label is positive. The F1-score of Normalization Alignment Tagging can reach 0.95\% on the development set. After recognizing words needing normalization, we run a statistical normalization method as described in Appendix A. This step can cover and predict the Normalization Alignment.

\paragraph{Null-Aligned Concept Tagging} For concepts that do not have alignment with input words, we define trigger words for those concepts and also use sequence tagging method. To be more specific, we first collect the dictionary of all Null-Aligned concepts in the training set and there are 184 different Null-Aligned concepts in total. The label of the input word is the class of Null-Aligned concept it triggers, or ``None'' if it triggers nothing.

For Null-Aligned concept, we define the concepts that it has a direct relation to as its trigger concepts and the aligned word of the trigger concept as the trigger word. For example, as shown in Figure~\ref{fig:camr}, the Null-Aligned concept ``Event'' has direct relation to concept ``\begin{CJK*}{UTF8}{gbsn}钱塘江大潮\end{CJK*}''. According to the alignment information of the concept, the trigger words of the concept ``Event'' are x1 and x2. Since a Null-Aligned concept could have multiple concepts it has a direct relation to, we tried using the first or the last of the concepts. The experimental result shows that using the last concept is more effective with a 0.03 F1 improvement.

There are nearly 5\% cases where the trigger concepts are all Null-Aligned concepts. Under such circumstances, we keep tracing back from the trigger concept until we reach the first concept with alignment and we regard this concept as the trigger concept.

\subsection{Relation-Prediction}

As shown in Figure~\ref{fig:models}, we design a RoBERTa-BiLSTM-Biaffine network to conduct relation prediction given the predicted concepts. All concepts are first split into characters before feeding into the RoBERTa model to extract hidden representations. After the RoBERTa model, all hidden states are fed into a one-layer BiLSTM network to better encode sequential information to the hidden states. At last, the \textbf{first hidden states of every two concepts} are fed into the biaffine network to get the relation between the two concepts. During training, we use Cross-Entropy as the loss function and use the average loss of $N\times N$ relations as the final loss, as described in Equation~\ref{eq:biaffine},

\begin{equation}
\begin{aligned}
    \text{Biaffine}(\mathbf{a},\mathbf{b})  & = [\mathbf{a};\mathbf{1}]\mathbf{W}[\mathbf{b};\mathbf{1}]^T, (\mathbf{a} \in \mathbb{R}^{1 \times d}, \mathbf{b} \in \mathbb{R}^{1 \times d},  \mathbf{W} \in \mathbb{R}^{(d+1)\times c \times(d+1)}) \\
    \text{Loss}&=\frac{1}{N^2}\sum_{i=1}^N \sum_{j=1}^N CE(\text{Biaffine}(\mathbf{h_i},\mathbf{h_j}), \hat{r}(\mathbf{h_i},\mathbf{h_j}))\\
    \text{Relation}_{a,b} & = \arg \max  \text{Biaffine}(\mathbf{a},\mathbf{b}) 
\end{aligned}
\label{eq:biaffine}
\end{equation}

where $d$ denotes the hidden size, $c$ denotes the number of relations, $N$ denotes the number of input concepts, $\hat{r}$ denotes the one-hot vector for the gold relation and $\mathbf{h_i}$ denotes the $i^{th}$ output hidden state from the BiLSTM network.

\paragraph{Relation-Alignment Prediction} On top of relation between concepts, another important feature of Chinese AMR is the relation alignment, which takes Chinese functional words' semantics into consideration in AMR graph. For example,  as shown in Figure~\ref{fig:camr}, functional word ``\begin{CJK*}{UTF8}{gbsn}被\end{CJK*}'' is aligned to the ``arg1'' relation between concepts ``\begin{CJK*}{UTF8}{gbsn}称为-01\end{CJK*}'' and ``Event''. In fact, in the input Chinese sentence, the word ``\begin{CJK*}{UTF8}{gbsn}被\end{CJK*}'' 
is the marker of relation ``arg1'' between concept ``\begin{CJK*}{UTF8}{gbsn}钱塘江大潮\end{CJK*}'' and ``\begin{CJK*}{UTF8}{gbsn}称为-01\end{CJK*}''.

\begin{figure}[t]
    \centering
    \includegraphics[width=0.8\linewidth]{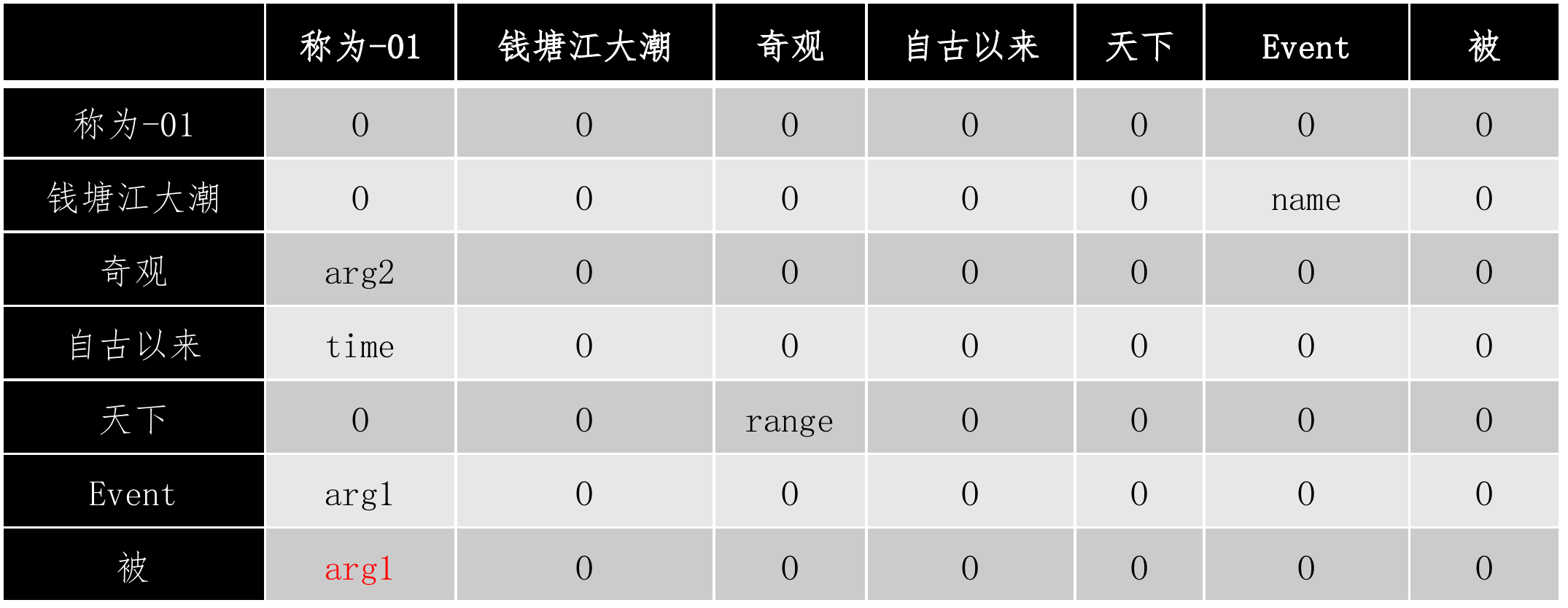}
    \caption{An example of Relation Classification label matrix. The inputs are from the gold concepts of sentence ``\begin{CJK*}{UTF8}{gbsn}钱塘江大潮被称为天下奇观\end{CJK*}''. Each column denotes the start node of a relation while each row denotes end node of a relation. ``O'' denotes there is no relation between the two nodes. Relation in red denotes the relation alignment for functional words.}
    \label{fig:relaton_example}
\end{figure}

We use the same model as relation prediction to align functional words with relations. To be more specific, concepts and functional words are both fed into the RoBERTa-BiLSTM-Biaffine network. As depicted in Figure~\ref{fig:relaton_example}, for any relation triples(concept1, concept2, relation), if the relation is aligned with functional word $w$, we create another triple(concept1, $w$, relation) for the model to predict. In this way, we can predict the relation and relation alignment simultaneously. After predicting all relations, if one concept is linked with one concept and one functional word with the same relation, the functional word will be aligned to the relation.

\subsection{Teacher Forcing in Training}

Since our method has two stages, during inference the Relation-Prediction model takes the output of Concept-Prediction as input. To stabilize and prevent error propagation during training, we adopt the Teacher Forcing method, where we use the gold concepts and relations as the input of the relation prediction model. However, error propagation still exists in the inference phase. We will discuss the error propagation situation of our system in section~\ref{sec:error_p}.

\section{Experiment}

\begin{table}[t]
    \centering
    \resizebox{0.4\textwidth}{!}{
    \begin{tabular}{lcc}
\toprule

Dataset Split & Sentences & Tokens \\

\midrule

Train & 16576 & 386234\\
Development & 1789 & 41822\\
Test & 1713 & 39228 \\
Blind Test & 1999 & 36940 \\

\bottomrule

\end{tabular}}
    \caption{The dataset description  of CAMRP-2022. Note that the train, development and test splits are directly from CAMR 2.0 dataset while the blind test split is from this evaluation.}
\label{tab:dataset}
\end{table}

\subsection{Dataset}
\label{sec:dataset}
The CAMRP-2022 evaluation uses the training, development, and test splits of the CAMR 2.0 dataset as its dataset and also involves an out-of-domain blind test set to measure the generalization performance of parsers. The statistics of the dataset are shown in Table~\ref{tab:dataset}. For concepts, there are 31941 different concepts in the training set. Among all concepts, there are 8443 different predicates that need to conduct word sense disambiguation and 184 Null-Aligned concepts. As for relations, there are 142 different relations and 841 relation alignment words. The top 5 most frequent relation alignment words are ``\begin{CJK*}{UTF8}{gbsn}的\end{CJK*}'',``\begin{CJK*}{UTF8}{gbsn}是\end{CJK*}'',``\begin{CJK*}{UTF8}{gbsn}和\end{CJK*}'',``\begin{CJK*}{UTF8}{gbsn}在\end{CJK*}'' and ``\begin{CJK*}{UTF8}{gbsn}对\end{CJK*}''.

\subsection{Model}

Both the Concept Tagging and Relation Classification model adopt the HIT-roberta-large\cite{cui-etal-2020-revisiting} pretrained model downloaded from HuggingFace model hub\footnote{https://huggingface.co/hfl/chinese-roberta-wwm-ext-large}. For Concept Tagging models, the output size of the tag classifier is 8, 2, 185 for Surface Tagging, Normalization Alignment Tagging, and Null-Aligned Concept Tagging individually and the dropout rate of the classifiers is 0.1 in all experiments. For the Relation Classification model, there is one BiLSTM layer and the hidden size is 4096. the dimension of Biaffine matrix is $4097\times142\times4097$. 

\subsection{Training Details}

We use Adam as the optimizer and conduct hyper-parameter searches on batch-size (from 10 to 100) and learning rate (from 1e-5 to 1e-4 ) in all models. The optimal hyper-parameters for each model are listed in Appendix B.  We train all models for 100 epochs with 1\% warmup steps and select the one with the best result on the development set as the final model.

\subsection{Results}
\label{sec:result}

\begin{table}[t]
    \centering
    \resizebox{0.7\textwidth}{!}{
    \begin{tabular}{lccc}
\toprule

Task(Dev) & Precision & Recall & F1 \\

\midrule

Surface Tagging & 0.918 & 0.944 & 0.931 \\
Normalization Aligment Tagging & 0.878  & 0.878 & 0.878 \\
Null-Aligned Concept Tagging & 0.708 & 0.679 & 0.693 \\
Relation Classification (With Gold Concepts) & 0.751 & 0.737 & 0.744  \\

\midrule

AlignSmatch & 0.778 & 0.766 & 0.768\\
\quad - Only Instance &0.830 & 0.833 &0.832\\
\quad - Only Attribute &0.928&0.954&0.941\\
\quad - Only Relation &0.614&0.556&0.583\\

\bottomrule

\\

\toprule
Task(Test) & Precision & Recall & F1 \\
\midrule

AlignSmatch & 0.786 &0.765 &0.776 \\
\quad - Only Instance & 0.834  & 0.840 & 0.837\\
\quad - Only Attribute& 0.932 & 0.959 & 0.945\\
\quad - Only Relation & 0.628 & 0.570  & 0.598\\

\bottomrule

\\

\toprule
Task(Blind Test) & Precision & Recall & F1 \\

\midrule

AlignSmatch& 0.715 & 0.696& 0.705\\
\quad - Only Instance & 0.768  & 0.775  & 0.772\\
\quad - Only Attribute & 0.866  & 0.901  & 0.883\\
\quad - Only Relation & 0.549  & 0.492  & 0.519\\

\bottomrule

 \end{tabular}}
    \caption{The fine-grained results of our model in CAMRP-2022. We report the overall and fine-grained AlignSmatch scores of our model on the development, test and blind test sets. We also report the results of each sub-task in the two-stage method on the development set.}
\label{tab:result}
\end{table}

As shown in Table~\ref{tab:result}, we list the results of our trained 2-stage AMR Parser on the development, test, and blind-test set of CAMRP-2022. For the development set, we list the concrete results of all different sub-tasks in two stages along with the overall and fine-grained AlignSmatch scores. 

\paragraph{Sub-task Results}

For the three sub-tasks of the Concept-Prediction stage, we can tell from Table~\ref{tab:result} that our model performs better in the Surface Tagging task with a 0.931 F1 score and Normalization Alignment Tagging task (0.878 F1) than in Null-Aligned Concept Tagging task (0.693 F1). It suggests that the model can better recognize concepts with alignment and there is a big performance drop when predicting concepts without alignment under the same sequence tagging framework. For the Relation Classification task, our model can reach 0.744 F1 when given gold concepts while only 0.583 F1 in inference when the concepts are generated by the Concept-Prediction stage instead of gold concepts. It reveals a train-inference discrepancy existing in the current method since the model might generate wrong concepts during the Concept Prediction stage in inference which would bias the Relation Prediction stage.

\paragraph{AlignSmatch Results}
As for the overall AlignSmatch scores, we can tell from the result of Development, Test and Blind-Test evaluations that there exists a domain shift. When looking at the fine-grained scores, the trend is consistent among three evaluation dataset that the performance of attribute or alignment prediction is better than instance prediction and far better than relation prediction. The trend indicates that the model generally outperforms in the first stage than in the second stage.

Moreover, for relation prediction, we can see that the recall is about 5 points lower than the precision in all experiments and the gap is much bigger than instance or attribute prediction. The reason is that in the relation classification model there exists a performance gap in relation prediction and relation alignment prediction. Compared to relation prediction, a lot more relation alignments are not predicted while the relation-only score in AlignSmatch takes both relation and relation alignment into account, which makes the recall score lower. In fact, if we preclude relation alignment prediction in the relation-only score, the gap between precision and recall will be reduced to 2 points. It hints to us that we need to pay more attention to the relation alignment prediction to improve the overall performance.

\section{Discussion}
In this section, we summarize some problems that need to be addressed to improve the performance of the Chinese AMR parser.

\subsection{Error Propagation in the Two-Stage Model}
\label{sec:error_p}

As pointed out in Section~\ref{sec:result}, there exists error propagation in the two-stage model. The direct evidence is that while the relation prediction could reach 0.744 F1 with gold concepts, this score drops to 0.583 when giving it the model predicted concepts.  Error propagation also exists in the Concept-Prediction stage since Normalization Alignment needs both the correct result from Surface Tagging and Normalization Alignment Tagging. 

\subsection{Class Imbalance Problem}
As pointed out in section~\ref{sec:dataset}, there exist severe class imbalance problems in both stages of the parsing task. As for the Concept-Prediction stage, the problem reflects the great differences in the distribution of different tags in the three tagging tasks, especially for the Null-Aligned Concept Tagging tasks. For the Relation-Prediction stage, a large portion of labels is ``None Relation'' as shown in Figure~\ref{fig:relaton_example}.

We have tried some techniques like using weighted loss that assigns greater weight to the minority classes to handle the class imbalance problem. While this can greatly reduce the time required for the model to converge, it does not improve the final performance when all epochs are finished. 
\subsection{Improving the Null-Aligned Concept Prediction Performance}
In our model, we use a trigger-based method to predict concepts without alignment. This method could cover nearly 95\% cases while the rest 5\% is neglected because they are mostly triggered by another Null-Aligned concept. Though we design methods to overcome the drawback by tracing back to the first aligned concept, the overall result of Null-Aligned Concept Prediction is still the lowest in the Concept-Prediction stage, which could lead to great bias for the next stage. A more natural method to predict those concepts might greatly improve this task.

\section{Conclusion}

In this paper, we provide a detailed description of the proposed two-stage Chinese AMR Parsing model which is the first to deal with the explicit word alignment problem for CAMRP-2022 evaluation. We also analyze the result and point out the limitation of the current method and some potential roads that might lead to improvement. Though straightforward, the method is far from perfect that it still calls for future exploration to reach a better result in the Chinese AMR Parsing task.

\bibliography{ccl2022}
\bibliographystyle{ccl}

\newpage

% \section*{Appendix A: Word Normalization Cases}
% \label{app:A}

% \section*{Appendix B: Alignment Cases}
% \label{app:B}

\section*{Appendix A: Statistical Word Normalization Method}
\label{app:A}

\begin{table}[h]
    \centering
    \resizebox{0.75\textwidth}{!}{
    \begin{tabular}{ll}
\toprule

Case Description  &Examples \\

\midrule

Word Sense Disambiguation  & \begin{CJK*}{UTF8}{gbsn}合作: 合作-01, 成: 成-01, 大: 大-01\end{CJK*}  \\
Special Concept Transform  & \begin{CJK*}{UTF8}{gbsn}不: - , 在: be-located-at-91\end{CJK*} \\
Number Normalization  & \begin{CJK*}{UTF8}{gbsn}第一: 1, 一万: 10000\end{CJK*} \\
Error Correction  & \begin{CJK*}{UTF8}{gbsn}JY: 精英, 暴光: 曝光-01, 惊荒: 惊慌-01\end{CJK*}  \\

\bottomrule

\end{tabular}}
    \caption{Main cases that need word normalization.}
\label{tab:norm cases}
\end{table}

\begin{CJK*}{UTF8}{gbsn}
As shown in Table~\ref{tab:norm cases}, we mainly summarize 4 cases that require word normalization. The Word Sense Disambiguation case denotes that the concept is required to select a word sense in terms of semantic. In Table~\ref{tab:norm cases}, "合作" is interpreted as the No.1 meaning in the dictionary, so "合作" should become "合作-01". The Special Concept Transform case is that many concepts with similar semantics in the corpus are replaced by special token. For example, "不" and "否" are replaced by "-". The Number Normalization case is that all concepts containing numbers need to be converted to Arabic numerals. The Error Correction case denotes that there may be errors in the sentence and we need to correct it according to the dictionary. \end{CJK*}

We design a set of word sense disambiguation rules to realize word normalization. As listed in Table~\ref{tab:norm cases}, there are 4 categories of concepts that need to be normalized. For the 3rd case, namely, concepts that need to be converted into numbers, we extract the numbers in a sentence using regular expressions and then convert the numbers into Arabic numerals. For the 4th case, we retrieve the most similar word in the dictionary which are provided based on the phonological and calligraphical code of Chinese characters to replace the wrong one. For the other cases, we directly use the concept that appears most frequently in the training set.

When we simply use the concepts that appear most frequently in the training set for all cases, the accuracy on the development set is 85.1. When we incorporated the rules mentioned above, the accuracy increase to 90.6, proving that the rules we designed can effectively handle a number of normalized concepts.

\section*{Appendix B: Optimal Hyper-Parameters for different models}
\label{app:B}

\begin{table}[h]
    \centering
    \resizebox{0.7\textwidth}{!}{
    \begin{tabular}{lcc}
\toprule

Model & Batch Size & Learning Rate \\

\midrule

Surface Tagging & 10  & 2e-5 \\
Normalization Tagging & 40  & 3e-5 \\
Null-Aligned Concept Tagging & 30  & 3e-5  \\
Relation Classification & 50  & 7e-5  \\

\bottomrule

\end{tabular}}
    \caption{The optimal hyper-parameters in each model.}
\label{tab:dataset}
\end{table}

\end{document}